\pdfoutput=1

\documentclass[11pt]{article}

\usepackage[preprint]{acl}

\usepackage{times}
\usepackage{latexsym}

\usepackage[T1]{fontenc}

\usepackage[utf8]{inputenc}

\usepackage{microtype}

\usepackage{inconsolata}

\usepackage{graphicx}
\usepackage{booktabs}
\usepackage{pifont}
\usepackage{amsthm,amsmath,amssymb}
\usepackage{multirow} 
\usepackage{tcolorbox}
\usepackage{hyperref}
\usepackage{enumitem}
\usepackage{multicol}
\usepackage{booktabs} 
\usepackage{fontawesome}
\usepackage{array}
\usepackage{colortbl} 
\usepackage{xcolor} 
\usepackage{subcaption}

\newcommand{\up}[1]{\textsubscript{\color{green!70!black}{↑#1}}}
\newcommand{\down}[1]{\textsubscript{\color{red!70!black}{↓#1}}}
\title{Precision over Diversity: High-Precision Reward Generalizes to Robust Instruction Following}

\author{
    \textbf{Yirong Zeng\textsuperscript{1*}},
  \textbf{Yufei Liu\textsuperscript{2*}},
  \textbf{Xiao Ding\textsuperscript{1}},
  \textbf{Yutai Hou\textsuperscript{3}},
 \textbf{Yuxian Wang\textsuperscript{3}},
 \textbf{Haonan Song\textsuperscript{3}}, \\
 \textbf{Wu Ning\textsuperscript{3}},
 \textbf{Dandan Tu \textsuperscript{3}},
 \textbf{Qixun Zhang\textsuperscript{2}},
 \textbf{Bibo Cai\textsuperscript{1}},
 \textbf{Yuxiang He},
 \textbf{Ting Liu\textsuperscript{1}}, 
\\
\\
 \textsuperscript{1}Harbin Institute of Technology, SCIR,
 \textsuperscript{2}Peking University, \\
 \textsuperscript{3}Huawei Technologies Co., Ltd,
\\
\small{
    \textsuperscript{*}Equal contribution, Work done during internship at Huawei.
  }
}

\begin{document}
\maketitle
\begin{abstract}
A central belief in scaling reinforcement learning with verifiable rewards for instruction following (IF) tasks is that, a diverse mixture of verifiable hard and unverifiable soft constraints is essential for generalizing to unseen instructions. 
In this work, we challenge this prevailing consensus through a systematic empirical investigation. 
Counter-intuitively, we find that models trained on hard-only constraints consistently outperform those trained on mixed datasets.
Extensive experiments reveal that reward precision, rather than constraint diversity, is the primary driver of effective alignment. 
The LLM judge suffers from a low recall rate in detecting false response, which leads to severe reward hacking, thereby undermining the benefits of diversity. 
Furthermore, analysis of the attention mechanism reveals that high-precision rewards develop a transferable meta-skill for IF.
Motivated by these insights, we propose a simple yet effective data-centric refinement strategy that prioritizes reward precision.
Evaluated on five benchmarks, our approach outperforms competitive baselines by 13.4\% in performance while achieving a 58\% reduction in training time,
maintaining strong generalization beyond instruction following.
Our findings advocate for a paradigm shift: moving away from the indiscriminate pursuit of data diversity toward high-precision rewards.

\end{abstract}


\section{Introduction}


Instruction Following (IF) serves as a primary metric for assessing model to follow the user's instruction constraints.
These constraints are generally classified into two categories: hard constraints, which are objectively enforceable through strict rules (e.g., word count), and soft constraints, which necessitate semantic interpretation (e.g., tone adjustment) ~\citep{zhou2023instruction}\footnote{In this work, we use the terms \textit{soft} and  \textit{semantic} constraints, \textit{hard} and \textit{rule} constraints interchangeably.}.
To achieve superior IF performance, the research community has recently turned to Reinforcement Learning with Verifiable Rewards (RLVR), spurred by its recent breakthroughs~\citep{guo2025deepseek,OpenAIGPT5}.
Building on this momentum, contemporary works increasingly use diverse constraints to extend RLVR.
The prevailing hypothesis posits that combining hard and soft constraints is key to achieving strong, generalizable performance~\citep{pyatkin2025generalizing,guo2025recast,peng2025verif}.

\begin{figure}[t]
    \centering
  \includegraphics[width=0.95\columnwidth]{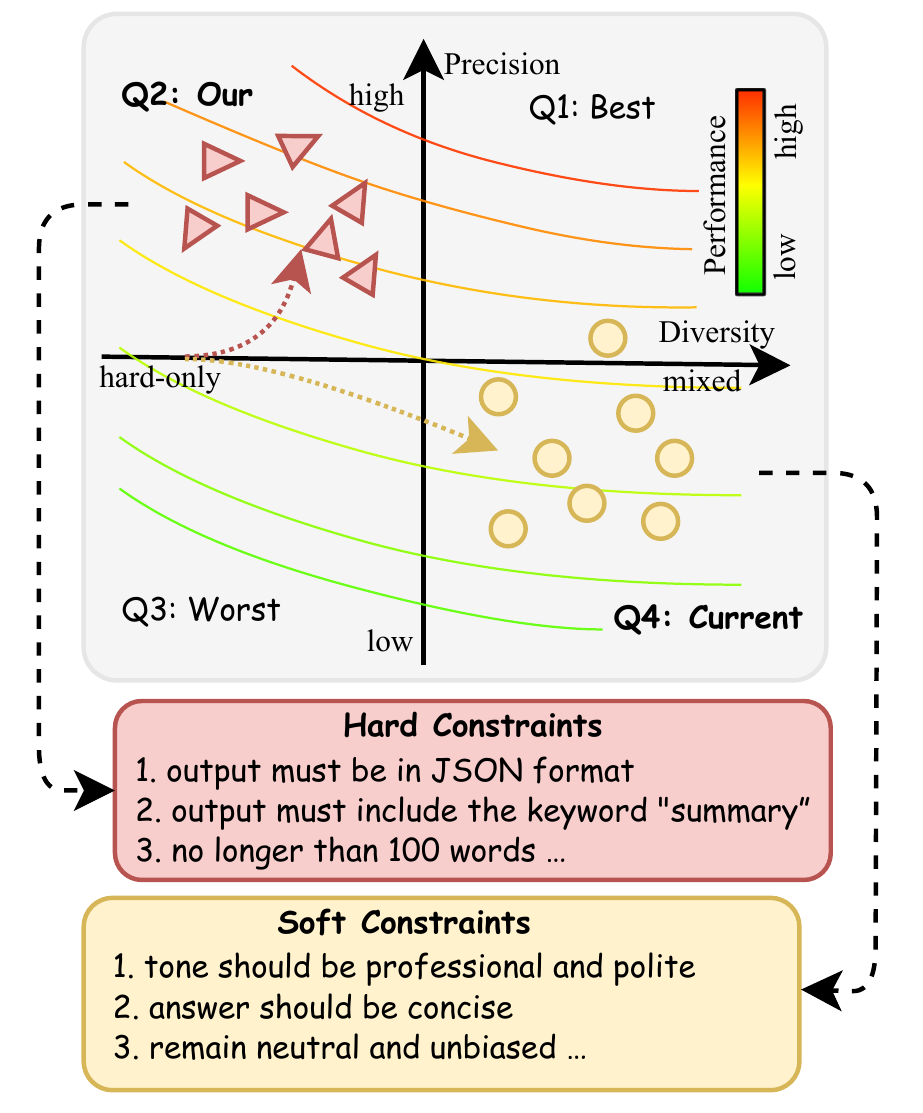}
  \caption{The impact of data distribution (reward precision and constraint diversity) on model generalization performance. 
  The contour lines represent model performance when trained under different data distributions.
  Counterintuitively, we find that high-precision rewards, even with limited diversity, achieve superior generalization compared to more diverse reward signals.
  }
  \label{fig:intro}
\end{figure}

In this work, we challenge this prevailing intuition.
Through systematic experiments,  we constructed datasets containing exclusively verifiable hard constraints (\textit{hard-only}), exclusively soft constraints (\textit{soft-only}), and a mixture of both to benchmark their impact on RLVR.
We observe a counter-intuitive phenomenon: models trained exclusively on verifiable hard constraints consistently outperform those trained on soft-only or diverse mixed datasets, even when evaluated on soft-only IF benchmarks.
This phenomenon remains robust across various model architectures and benchmarks.
It suggests that the view (equating broader constraint diversity with better IF generalization) calls for a fundamental re-evaluation.

To investigate the origins of this performance disparity, we analyze the reward reliability of both constraints.
The result reveals that: 
Hard constraints utilize rule-based verifiers that serve as a near-perfect ground truth; 
Soft constraints rely on LLM-based judges, which suffer from significantly lower reward precision. 
We also observe that the LLM judge exhibits a low recall rate in detecting false responses, reflecting its susceptibility to reward hacking (e.g., it often fails to penalize outputs that violate constraints).
Additionally, we design experiments to quantify the relationship between reward precision and diversity.
Its results show that lower reward precision leads to a marked performance drop, even with high diversity, whereas increasing diversity alone yields little to no benefit.
As illustrated in Figure \ref{fig:intro}, reward precision, not diversity, is the primary driver of effective IF.

Furthermore, we explain the generalization puzzle: why training on narrow hard constraints transfer to broad semantic instructions? 
Our analysis of attention mechanisms reveals that high-precision RLVR develops a transferable meta-skill rather than merely memorizing specific rules.
By internalizing IF capabilities, the model exhibits sparse yet precise attention patterns on constraint segments while maintaining sufficient attention on query segment to support generalization. 
In other words, the model has internalized instruction-following capability. it learns how to follow instructions, not just what to follow.

Motivated by this insight, we propose 
a simple yet effective data-centric refinement strategy that prioritizes reward precision over constraint diversity. 
It consists of:
(1) filtering out practically unsatisfiable constraints to improve reward precision, and
(2) limiting each training instance to at most one soft constraint to mitigate reward hacking by the LLM-as-a-judge.
Evaluated on five IF benchmarks (e.g., IFEval and CFBench), our method achieves an average 13.4\% performance gain over baselines, surpassing the mixed-constraint model by 7.5\%. 
Crucially, this improvement is achieved while preserving general capabilities and generalization, all with a 58\% reduction in training time.

\begin{table*}[th]
    \centering
    \small
    \begin{tabular}{l | ccc|cc|c}
        \toprule
         Model & \ding{168}IFEval & \ding{168}Multi-IF & \ding{168}IFBench & \ding{171}CFBench & \ding{171}FollowBench & \textbf{Average} \\
        \midrule \rowcolor{lightgray!40}
        Qwen2.5-7B-inst & 72.46 & 51.05 & 28.91 & 44.00 & 61.40 & 51.56 \\ 
        \hspace{0.2cm} w/ \textit{Hard-only} & 80.78 & 58.89 & 31.63 & 49.00 & 68.96 & \textbf{57.85} \\
        \hspace{0.2cm} w/ \textit{Soft-only} & 77.82 & 54.44 & 27.89 & 47.00 & 68.73 & 55.18 \\
        \hspace{0.2cm} w/ \textit{Mix} & 78.37 & 58.25 & 29.59 & 51.00 & 69.07 & 57.26 \\
        \midrule \rowcolor{lightgray!40}
        Qwen2.5-32B-inst & 81.70 & 64.45 & 33.67 & 57.00 & 73.06 & 61.98 \\ 
        \hspace{0.2cm} w/ \textit{Hard-only} & 84.10 & 68.59 & 35.71 & 60.00 & 75.12 & 64.70 \\
        \hspace{0.2cm} w/ \textit{Soft-only} & 82.99 & 66.04 & 30.95 & 58.00 & 74.19 & 62.43 \\
        \hspace{0.2cm} w/ \textit{Mix} & 83.55 & 68.87 & 37.75 & 60.00 & 74.99 & \textbf{65.03} \\
        \midrule \rowcolor{lightgray!40}
        Qwen3-8B & 85.77 & 70.36 & 24.48 & 55.00 & 67.28 & 60.58 \\ 
        \hspace{0.2cm} w/ \textit{Hard-only} & 88.54 & 73.37 & 25.17 & 57.00 & 67.79 & \textbf{62.37} \\
        \hspace{0.2cm} w/ \textit{Soft-only} & 86.32 & 70.97 & 27.21 & 56.00 & 67.21 & 61.54 \\
        \hspace{0.2cm} w/ \textit{Mix} & 86.51 & 72.73 & 25.85 & 57.00 & 67.70 & 61.96 \\
        \midrule \rowcolor{lightgray!40}
        Llama3.2-3B-inst & 74.12 & 40.99 & 20.74 & 17.00 & 50.40 & 40.65 \\ 
        \hspace{0.2cm} w/ \textit{Hard-only} & 78.00 & 53.00 & 25.85 & 22.00 & 53.65 & 46.50 \\
        \hspace{0.2cm} w/ \textit{Soft-only} & 74.86 & 44.75 & 24.82 & 22.00 & 54.11 & 44.11 \\
        \hspace{0.2cm} w/ \textit{Mix} & 79.30 & 51.31 & 24.48 & 25.00 & 53.61 & \textbf{46.74} \\
        \bottomrule
    \end{tabular}
    \caption{Evaluation results on five benchmarks. \ding{168} and \ding{171} denote hard-only constraints and mixed constraints (including soft constraints) benchmarks, respectively.
    \textit{Hard-only} models consistently outperform \textit{Soft-only} models across various base model configurations.
    }
    \label{tab:pre-exp1}
\end{table*}

\section{The Effectiveness of Verifiable Constraints}
\label{sec:analysis}
In this section, we systematically investigate the impact of constraint type on RLVR training,
presenting a counter-intuitive phenomenon where training on hard-only constraints generalizes better than training on diverse constraints. 

\subsection{Experimental Setup}
\textbf{Datasets and Benchmarks.} 
In model training, we use the \textit{VerInstruct} train dataset from ~\citet{peng2025verif}, which contains 22,000 instances: 77.7\% with soft constraints (reward by LLM-as-a-judge) and 22.3\% with hard constraints (reward by code-based rules).
We evaluate the models on some representative IF benchmarks, including IFEval ~\citep{zhou2023instruction}, Multi-IF ~\citep{he2024multi}, and IFBench ~\citep{pyatkin2025generalizing}, all of which focus exclusively on verifiable hard constraints.
FollowBench ~\citep{jiang2024followbench} and CFBench ~\citep{zhang2025cfbench} cover a comprehensive range of mixed constraint types, including soft constraints.
The benchmark details are provided in Appendix \S\ref{sec:benchmark}.
For evaluation, we use the Instruction Satisfaction Rate (ISR; ~\citealp{zhang2025cfbench}) as the metric, which measures the model’s ability to satisfy all constraints within a given query.

\textbf{Models and Training.} 
We select Qwen2.5-7B-Instruct ~\citep{qwen2025qwen25} as our base model, and Qwen3-32B as the reward model for soft constraints.
We employ the standard GRPO ~\citep{guo2025deepseek} algorithm for RLVR training of the base model. 
The training datasets are categorized into three subsets for comparison: \textit{hard-only}, \textit{soft-only}, and the raw \textit{mixed} constraint data.
For completeness, we present the detailed formulation of GRPO and more training in Appendix \S\ref{sec:exp_details}.

\subsection{Results}
\label{sec:pre_res}

\textbf{Performance Comparison.} 
We train models separately on the \textit{Hard-only}, \textit{Soft-only}, and \textit{Mix} datasets and evaluate them on the combined benchmark. 
As illustrated in Table \ref{tab:pre-exp1}, the model trained on \textit{Hard-only} data consistently outperforms the \textit{Soft-only} model, performs on par with the \textit{Mix} variants, surprisingly excelling even on semantic soft IF tasks (e.g., CFBench and FollowBench). 
For example, the \textit{Hard-only} model improves by 2.9\% on the in-distribution IFEval benchmark and by 2.0\% on the out-of-distribution CFBench, compared to the \textit{Soft-only} model.
Notably, this improvement is achieved using only 22.3\% of the training data, which underscores the surprisingly strong robustness of hard-only constraint training.
This finding contradicts the common intuition that a diverse mixture of constraints is necessary for broad generalization. 

\textbf{Robustness Check.} 
To ensure the validity of the above observation, we scale our experiments across different model architectures (e.g., Qwen2.5 ~\citep{qwen2025qwen25}, Qwen3 ~\citep{yang2025qwen3}, Llama3.2 ~\citep{dubey2024llama}) and sizes (3B, 7B, 8B, 32B). 
As shown in Table \ref{tab:pre-exp1},
the dominance of hard constraints remains robust across all settings, suggesting that this is a fundamental property of current RLVR paradigms rather than a model-specific artifact.

\begin{figure}[th]
    \centering
  \includegraphics[width=0.8\columnwidth]{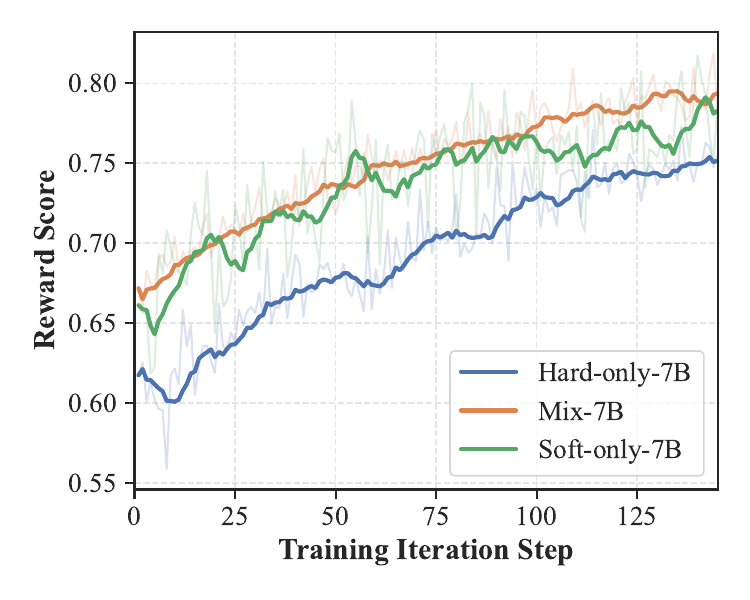}
  \caption{The model's training curves in hard-only, soft-only and mixed constraint datasets, respectively. 
  The \textit{Soft-only} model achieves higher reward scores than the \textit{Hard-only} model.
  }
  \label{fig:training_curves}
\end{figure}

\textbf{Visualization}. We visualize the training curves in Figure~\ref{fig:training_curves}.
The results reveal a notable discrepancy between the models: compared to the hard-only model, the soft-only model achieves higher reward scores by $\sim4\%$ during training but yields lower test performance by $2\%$.
\textit{This suggests that soft-only models are exploiting biases in LLM-as-a-judge to game the reward scores, rather than genuinely mastering constraint adherence. }
In contrast, rewards derived from code-based rules provide a more faithful and reliable measure of actual model capability.
This fundamental divergence between reward metrics calls for a critical re-evaluation of current reward engineering strategies, particularly those reliant on subjective LLM feedback.

\section{Diagnosing the Disparity: Reward Engineering }
In this section, we diagnose the problem by: measuring the reward strategies reliability, analyzing the failure mechanisms of LLM judge, and decoupling reward precision and diversity.

\begin{table}[t]
    \centering
    \small
    \begin{tabular}{l|cc|cc}
        \toprule
        \multirow{2}{*}{\textbf{Reward Model}} & \multicolumn{2}{c|}{\cellcolor{lightgray!40}{{Hard Constraint} } } & \multicolumn{2}{c}{\cellcolor{lightgray!40}{Soft Constraint}} \\
        & \textbf{Prec.} & \textbf{Rec.\textsubscript{}} & \textbf{Prec.} & \textbf{Rec.\textsubscript{}} \\
        \midrule
        Rule Checker & \textbf{96.0} & \textbf{81.2} & -- & -- \\
        \midrule
        Gemini-2.5-pro & 86.0 & 65.7 & \textbf{86.3} & \textbf{63.8}  \\
            \hspace{0.1cm} \textit{w/ pointwise} & 85.8 & 69.2  & 88.7 & 74.5 \\
        Qwen-3-32B & 76.5 & 30.6 & 74.5 & 20.9\textsubscript{\textcolor{red!70!black}{↓43.0}}  \\
            \hspace{0.1cm} \textit{w/ pointwise} & 82.6\textsubscript{\textcolor{green!70!black}{↑6.1}}  & 59.3 & 83.5 & 54.7  \\
        QwQ-32B & 80.7 & 44.1 & 78.2 & 33.1  \\
            \hspace{0.1cm} \textit{w/ pointwise} & 83.8 & 61.7 & 83.9 & 57.5  \\
        Qwen-2.5-32B &  {71.5} & {9.2} & {71.8} & {12.3}  \\
            \hspace{0.1cm} \textit{w/ pointwise} & 74.5 & 29.6  & 75.8 & 39.3 \\
        \bottomrule
    \end{tabular}
    \caption{Overall reward reliability evaluation results. \textbf{Prec.} denotes reward precision; \textbf{Rec.\textsubscript{}} represents the recall of false responses, reflecting error-detection capability;
    \textit{w/ pointwise} denotes evaluating each constraint individually in multi-constraint instructions, reflecting the model’s ability to judge single constraints.
    }
    \label{tab:llm_reward}
\end{table}

\begin{figure}[t]
    \centering
    \includegraphics[width=0.99\columnwidth]{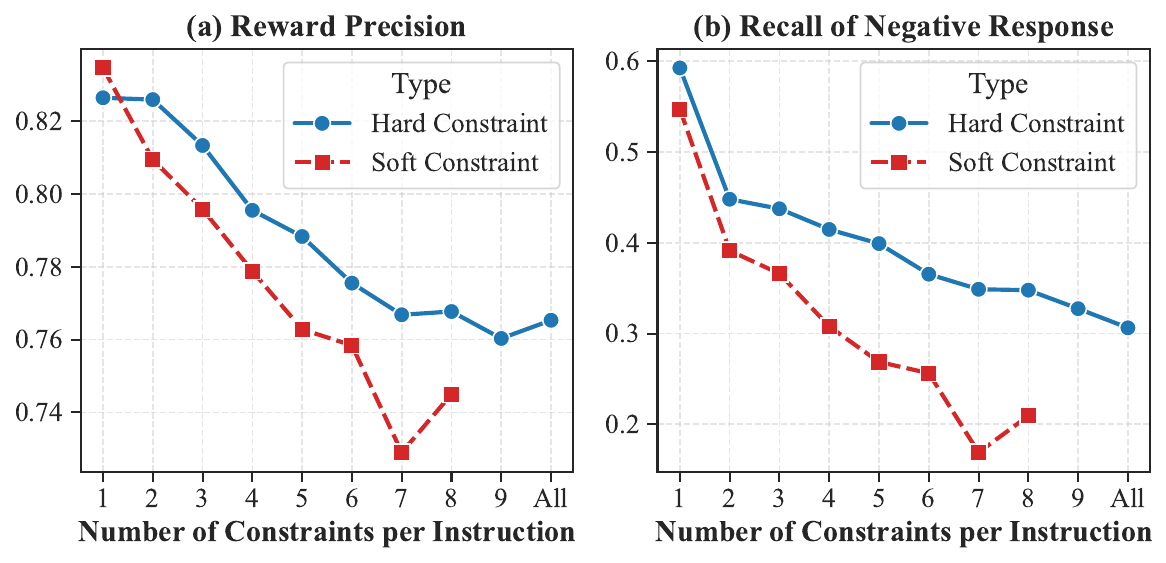}
    \caption{The reward reliability of LLM-as-a-judge under increasing the constraints. 
    It shows a clear degradation in performance emerges with increasing constraint diversity.
    }
    \label{fig:num_const}
\end{figure}

\subsection{Quantifying Reward Reliability}
\label{sec:reward_hack}
We evaluate the reward signals precision derived from the two types of constraints.
Specifically, for instances involving hard constraints, we sample responses from base model and assess them using both the rule checker and LLM-based reward model. 
For instances with soft constraints, we evaluate responses generated by the base model using the LLM-based reward.
We also establish ground-truth labels by human experts, details in Appendix \ref{sec:human_label}.
This allows us to quantify how well the reward module aligns with human evaluations.

Detailed evaluation results are presented in Table~\ref{tab:llm_reward}. 
For hard constraints, rule-based verifiers serve as a near-perfect ground truth, achieving a 96\% agreement rate with human judgments in reward precision (\textbf{Prec.}). 
In contrast, LLM rewards exhibit a significant performance gap compared to rule-based checkers, e.g., the widely used Qwen-3-32B lags behind by 19.5\%.
Even the most advanced closed-source Gemini-2.5-Pro\footnote{API-accessed models are impractical as reward models due to API rate limits.}  still underperforms relative to code-based verifiers by 9.7\%.
This suggests that current LLMs are not yet suitable for direct use as primary rewards. 
Similar conclusions are drawn from the soft constraints evaluation.
Furthermore, we observe that stronger foundation models achieve higher reward precision, implying that enhancing foundation model capabilities remains a viable path for improving reward precision.

To investigate the failure mechanisms of LLM rewards, we present an extended analysis in Table~\ref{tab:llm_reward}.
First, we evaluate point-wise verification (assessing each constraint individually) as an alternative to the previous batch judgment (assessing all constraints simultaneously) per instruction.
The results show that \textit{pointwise} judgment consistently outperforms batch judgment across open-sourced models, e.g., Qwen-3-32B achieves a 6.1\% improvement on hard constraints under \textit{pointwise}. 
This suggests that when models assess multiple constraints simultaneously, some form of bias interferes with reward precision.
Second, we report the reward \textbf{Rec.\textsubscript{}} to quantify the model's ability to detect non-compliant responses.
Across all LLM rewards, \textbf{Rec.\textsubscript{}} is consistently 20\%+ lower than \textbf{Prec}. 
Notably, open-sourced models exhibit a substantial gap compared to Gemini-2.5-pro, e.g., Qwen-3-32B showing a 43\% lower recall in soft constraint evaluations.
This highlights a specific failure mode of LLM rewards: an inability to penalize violations, indicating limited error-detection capability when multiple constraints are involved.
Such partial-credit noise enables the model to game the LLM judge and prevents the policy model from learning strict adherence boundaries.
We characterize this phenomenon as a form of inherent reward hacking within LLMs.


To quantify the impact of LLM reward hacking, we evaluate how the number of constraints per instruction affects LLM judge.
Figure~\ref{fig:num_const} illustrates the reward precision and negative recall across varying constraint counts. 
It reveals a pronounced downward trend in performance as complexity increases. 
While the model exhibits decent performance under a single constraint, it exhibits a substantial decline in reward precision and recall when adding a second constraint ($1\rightarrow2$). 
For instance, in soft constraint evaluations, \textbf{Prec} decreases by 2.5\%, whereas \textbf{Rec.\textsubscript{}} decreases by 5.6\% when adding a second constraint. 
This degradation suggests that LLM-derived rewards suffer severely from multi-constraint complexity, indicating that current LLMs are not yet robust enough to handle such complex scenarios.

\subsection{Decoupling Precision and Diversity}
To quantify the relative importance of reward precision versus constraint diversity, we design two controlled experiments across different training sets.

\begin{figure}[t]
    \centering
  \includegraphics[width=0.99\linewidth]{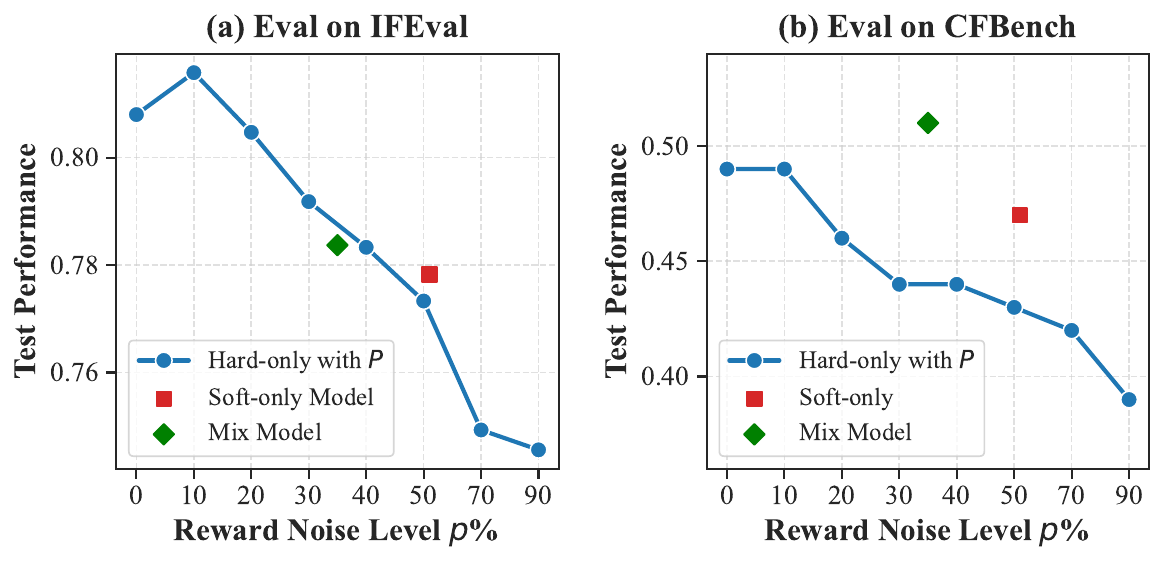}
  \caption{Impact of reward precision with random noise.
  It shows reward noise is the critical factor of test performance. 
  (a) Under identical noise levels, hard-only and soft-only model yield near performance. 
  (b) Notably, with low reward noise, hard-only model also fail to develop generalization capabilities.
  }
  \label{fig:prec_a}
\end{figure}

\textbf{Impact of Reward Precision.} 
 To this end,
 we simulate low-precision environments by injecting random noise into reward verifiers during hard-only training.
 Concretely, with probability $p$, we forcibly set the reward signal to 1, indicating that all constraints are satisfied, regardless of its original value (0 or 1).
This setup allows us to characterize the impact of noise in error-detection capability (e.g., \textbf{Rec.\textsubscript{}}).
 As shown in Figure~\ref{fig:prec_a}, we observe an overall decline in model performance as the noise level increases.
This expected trend underscores the significant impact of reward precision on model effectiveness.
Unexpectedly, we find that when $p \leq 10\%$, the model shows no performance degradation, even improves slightly on IFEval. 
 We attribute this robustness to two factors: 
 (1) low-frequency noise often preserves the relative rankings essential for GRPO’s group-based optimization, 
 and (2) low reward noise encourages the policy model to explore novel trajectories beyond the known solution paths~\citep{yue2025does}.

 We further estimated the effective noise level $p$ for the Soft-only model and the Mix model based on the \textbf{Rec.\textsubscript{}} gap relative to Rule Checker (Table \ref{tab:llm_reward}). 
 On the in-distribution IFEval benchmark, Hard-only, Soft-only, and Mix models yield nearly identical performance (e.g., $\Delta<0.5\%$ at $p=50$) at equivalent $p$ levels. 
 This suggests that reward precision, rather than diversity, is the primary determinant of model performances. 
 Conversely, on the out-of-distribution (OOD) CFBench, Hard-only model significantly outperforms the Soft-only and Mix baseline. 
 This indicates that under low-precision reward conditions, hard-only training fails to foster inherent generalization for OOD scenarios, similar to soft-only training, making in-distribution alignment a more effective strategy in such regimes.

\begin{figure}[th]
    \centering
  \includegraphics[width=0.9\linewidth]{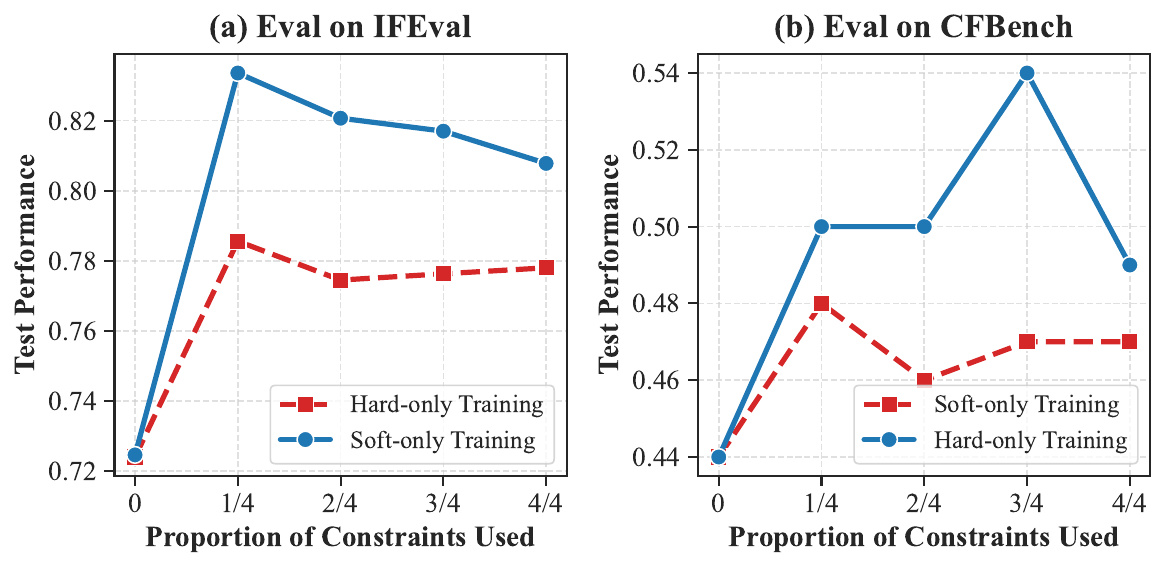}
  \caption{Impact of constraint diversity with using limited constraints.
  It can not observe a consistent improvement as the number of constraints grow.
  }
  \label{fig:prec_b}
\end{figure}


\textbf{Impact of Constraint Diversity.}
To investigate this, we progressively increase the number of constraints per instruction during training, considering both soft-only and hard-only reward settings.
As shown in Figure~\ref{fig:prec_b}, we do not observe a consistent improvement in test performance as the number of constraints grows.
Specifically, performance exhibits a noticeable gain when using $1/4$ constraints per instruction, but further increases lead to a slight decline.
This suggests that, regardless of whether rules or LLMs are used as a reward, pursuing excessive constraint diversity does not yield the anticipated benefits.
Once a minimal threshold of diversity is reached, adding more constraints results in diminishing or even negligible returns.

Therefore, combining these findings, we conclude that \textit{Reward Precision} significantly outweighs \textit{Reward Diversity} in RLVR for IF. 
The benefits of constraint diversity in train data are entirely outweighed by the negative impact of noisy rewards from the LLM judge.

\begin{figure}[th]
    \small
    \begin{subfigure}[t]{0.99\linewidth}
    \centering
        \includegraphics[width=0.8\linewidth]{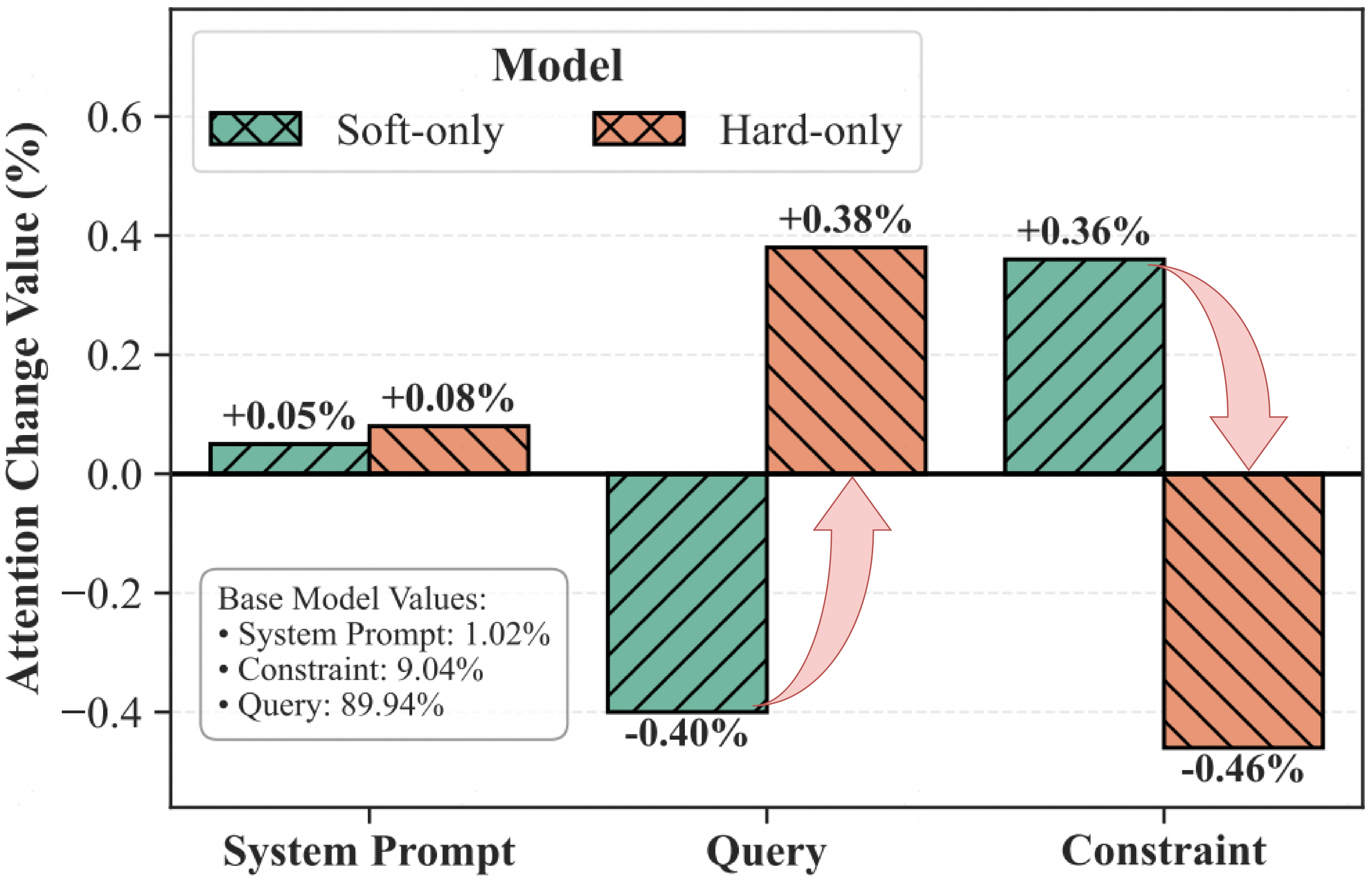} 
        \caption{Attention shift relative to base model on IFEval.}
        \label{fig:exp1a}
    \end{subfigure}
    \hfill
    \begin{subfigure}[t]{0.99\linewidth}
        \centering
        \includegraphics[width=0.8\linewidth]{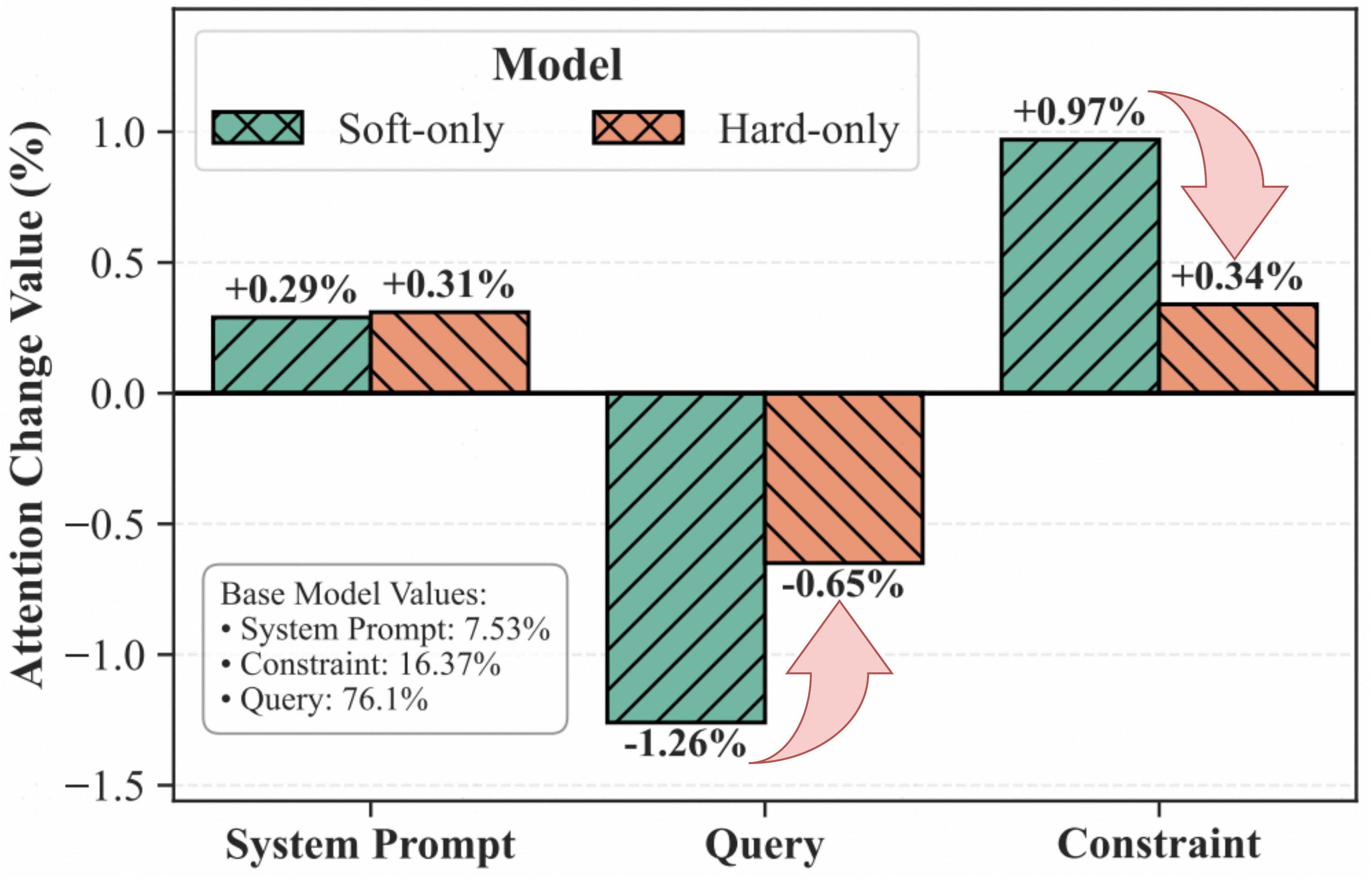}
        \caption{Attention shift relative to base model on CFBench.}
        \label{fig:exp1b}
    \end{subfigure}
  \caption{The attention weight distribution shift  relative to base Model.
  The soft-only model reduces attention to the query segment and instead places greater emphasis on the constraint, whereas the hard-only model exhibits the opposite trend.
  }
  \label{fig:att_m}
\end{figure}

\begin{figure}[t]
    \centering
  \includegraphics[width=0.99\linewidth]{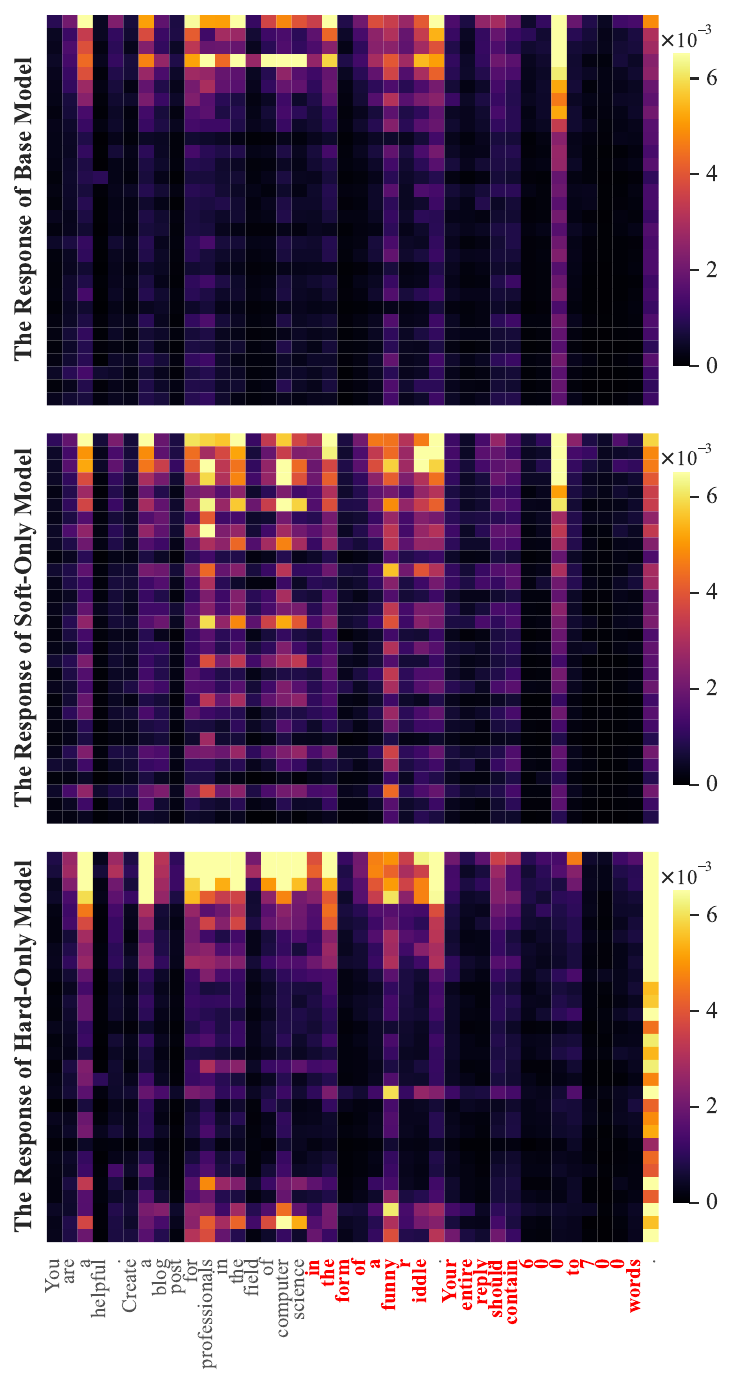}
  \caption{An attention heatmap analysis for a sample from IFEval. The \textcolor{red}{red text} denotes constraint segment.
  All models attend to the \textcolor{red}{funny riddle} region; and the base and soft-only models show more attention on the numbers \textcolor{red}{600 to 700} than the hard-only model.
  }
  \label{fig:att}
\end{figure}

\section{Mechanism Analysis: Why Does Hard-only Training Generalize?}
To shed light on why training on narrow hard constraints generalizes better than broad soft constraints, we analyze the shifts in attention patterns across the Base, \textit{soft-only}, and \textit{hard-only} models. 

Concretely, we quantify the model’s attention weights allocated to different prompt segments (i.e., system prompt, query, and constraint), and report the average weights across all instances (results illustrated in Figure~\ref{fig:att_m}).
On IFEval, we observe that the Soft-only model shifts its focus, decreasing its attention to the raw query by 0.4 while increasing its attention to constraints by 0.36. 
In contrast, the hard-only model increases attention to the query by 0.38 while reducing focus on constraints by 0.46.
Similar trends are observed on CFBench.
Given that the soft-only model underperforms the hard-only model, we conclude that:
(1) The soft-only model improves IF performance by allocating more attention to the constraint text, but at the cost of reduced attention to the raw query, leading to degraded generalization.
(2) In contrast, the hard-only model enhances IF without increasing attention to the constraint, suggesting that it has internalized IF capability; the freed-up attention capacity is instead redirected to the raw query segment, thereby improving generalization.
Therefore, this finding demonstrates that, after training with high-precision rewards, the model has internalized IF capability and acquired fundamental reasoning skills, thereby reducing the need for excessive attention to the constraint segment.


Figure~\ref{fig:att} provides a qualitative visualization of the attention maps for a randomly selected example.
It demonstrates that the hard-only model exhibits sparse yet precise attention patterns: by increasing attention efficiency, it reduces the average attention allocated to the constraint segment while preserving sufficient focus on the original user query.
In contrast, both the base and soft-only models display diffuse and inefficient attention distributions, which waste attentional resources and consequently limit their generalization capability.

Therefore, hard constraints serve as a {high-precision proxy},
internalizing the IF capacity.
The model acquires a meta-skill that naturally transfers to unverifiable soft tasks, effectively learning {how to follow} rather than just {what to follow}.


\begin{table*}[t]
    \centering
    \small
    \begin{tabular}{l|ccc|cc|c}
        \toprule
         Model & \ding{168}IFEval & \ding{168}Multi-IF & \ding{168}IFBench & \ding{171}CFBench & \ding{171}FollowBench & \textbf{Average} \\
        \midrule 
        Gemini-3.0-Pro & 95.56 & 83.10 & 68.70 & 70.00 & 84.00 & 80.27 \\ 
        Claude-Sonnet-4.5 & 91.13 & 81.57 & 44.55 & 63.00 & 79.46 & 71.94 \\
        DeepSeek-V3.2 & 91.87 & 74.14 & 45.91 & 67.00 & 79.76 & 71.74 \\
        Qwen3-8B & 85.77 & 70.36 & 24.48 & 55.00 & 67.28 & 60.58 \\
        Qwen3-32B & 87.06 & 70.70 & 25.17 & 64.00 & 69.61 & 63.31 \\
        \midrule
        IF-RLVR ~\citep{pyatkin2025generalizing} & 87.80 & --- & 53.70 & --- & --- & 70.75 \\
        RECAST ~\citep{guo2025recast} & 74.01 & --- & --- & --- & 63.23 & 68.62 \\
        Qwen-IF ~\citep{ren2025instructions} & 78.90 & --- & --- & 52.00 & 63.80 & 64.90 \\ 
        \midrule \rowcolor{lightgray!40}
        Qwen2.5-7B-inst & 72.46 & 51.05 & 28.91 & 44.00 & 61.40 & 51.56 \\ 
        \hspace{0.2cm} w/ \textit{Hard-only} & 80.78 & 58.89 & 31.63 & 49.00 & 68.96 & 57.85 \\
        \hspace{0.2cm} w/ \textit{Soft-only} & 77.82 & 54.44 & 27.89 & 47.00 & 68.73 & 55.18 \\
        \hspace{0.2cm} w/ \textit{Mix} & 78.37 & 58.25 & 29.59 & 51.00 & 69.07 & 57.26 \\
        \rowcolor{blue!10} \hspace{0.2cm} w/ HPPT-7B & 87.25\up{15.8} & 68.70\up{17.7} & 40.13\up{12.2} & 57.00\up{13} & 70.88\up{9.4} & 64.79\up{13.4} \\
        \midrule \rowcolor{lightgray!40}
        Qwen2.5-32B-inst & 81.70 & 64.45 & 33.67 & 57.00 & 73.06 & 61.98 \\ 
        \hspace{0.2cm} w/ \textit{Hard-only} & 84.10 & 68.59 & 35.71 & 60.00 & 75.12 & 64.70 \\
        \hspace{0.2cm} w/ \textit{Soft-only} & 82.99 & 66.04 & 30.95 & 58.00 & 74.19 & 62.43 \\
        \hspace{0.2cm} w/ \textit{Mix} & 83.55 & 68.87 & 37.75 & 60.00 & 74.99 & 65.03 \\
        \rowcolor{blue!10} \hspace{0.2cm} w/ HPPT-32B & 88.72\up{7.0} & 76.23\up{11.8} & 43.87\up{10.2} & 65.00\up{8} & 78.95\up{5.9} & 70.55\up{8.6} \\
        \bottomrule
    \end{tabular}
    \caption{Evaluation results on the five IF benchmarks. 
    Our HPPT models consistently outperform the base model and their variants.
    }
    \label{tab:final_exp}
\end{table*}

\section{Methods \& Experiments}
\label{sec:method}
We propose a simple data-centric refinement strategy to enable \textit{high-precision proxy training}. 
This strategy prioritizes reward signal reliability over constraint diversity through two key steps:

\textbf{(1) Denoising via learnability filtering.}
To eliminate practically unsatisfiable constraints that reduce precision by introducing noise, we implement a \textit{pilot training} phase. 
Formally, let $\mathcal{D}$ be the initial dataset. We train the policy $\pi_\theta$ on $\mathcal{D}$ for a short duration (e.g., 5 epochs) and track the reward trajectory $r(x, y)$ for each sample $(x, y)$. 
We then prune the dataset to create a high-precision subset $\mathcal{D}_{clean}$:
\begin{equation}
    \small
    \mathcal{D}_{clean} = \{ (x) \in \mathcal{D} \mid \exists t \in [1, T], r(x, y_t) > 0 \}
\end{equation}
where $T$ denotes the pilot training epochs.

\textbf{(2) Mitigating hacking via constraint simplification.}
Building on the empirical insights from Section \ref{sec:reward_hack}, the LLM-as-a-judge is prone to reward hacking when facing complex, multi-constraint instructions. 
To mitigate this, we strictly limit each training instance to contain \textbf{at most one soft constraint}. 
This simplification reduces cognitive load on the verifier, minimizing false negatives and ensuring that the reward signal remains precise and trustworthy.

\begin{table}[t]
    \centering
    \small
    \begin{tabular}{l|ccc}
        \toprule
        \multirow{2}{*}{\textbf{Model}} & \textbf{Math} & \textbf{Knowledge} & \textbf{Writing} \\
         & {\scriptsize GSM8K} & {\scriptsize MMLU} & {\scriptsize WritingBench} \\
        \midrule \rowcolor{lightgray!40}
        Qwen2.5-7B & 92.1 & 74.5 & 5.9 \\
        IF-RLVR  & 15.3\down{76.8} & 59.4\down{15.1} & - \\
        Qwen-IF & - & - & 5.8\down{0.1} \\
        \textbf{HPPT-7B} & \textbf{92.3}\up{0.2} & \textbf{73.1}\down{1.4} & \textbf{6.0}\up{0.1} \\
        
        \midrule \rowcolor{lightgray!40}
        Qwen2.5-32B & 95.4 & 78.0 & 6.1 \\
        \textbf{HPPT-32B} & \textbf{95.8}\up{0.4} & \textbf{80.9}\up{2.9} & \textbf{6.4}\up{0.3} \\
        \bottomrule
    \end{tabular}
    \caption{Evaluation of general capabilities, compared with RLVR-trained baselines. }
    \label{tab:general}
\end{table}

\textbf{Instruction-Following Performance.} 
The results are presented in Table~\ref{tab:final_exp}.  
For baseline comparisons, we report the performance of leading LLMs as well as recently proposed RLVR-trained models derived from the same base architecture.\footnote{As the training data and model checkpoints are not publicly available, we report only previously published results.}  
A detailed description of these baselines is provided in Appendix~\S\ref{sec:baseline}.
As shown in Table~\ref{tab:final_exp}, our HPPT models consistently outperform the base model, achieving average improvements of 13.23\% for the 7B variant and 8.57\% for the 32B variant.  
Moreover, under identical training settings, HPPT-7B surpasses the Mix-7B Model by 7.53\% despite using fewer constraint types.
Notably, HPPT-32B achieves performance within 1.2\% of top-tier models such as DeepSeek and Claude.  
These results demonstrate that high-precision proxy training can serve as a competitive alternative to LLM alignment approaches.
We present an ablation study of HPPT in Appendix~\ref{sec:ablation}.

\textbf{General Capabilities Performance.} 
Beyond evaluating IF capability, we also assess the model's general proficiency across three dimensions: 
mathematics using GSM8K ~\cite{cobbe2021training}, general knowledge using MMLU ~\cite{hendrycks2020measuring}, and writing ability using WritingBench ~\cite{wu2025writing}.
As shown in Table \ref{tab:general}, previous RLVR-trained models exhibit a performance decline in these general tasks (e.g., IF-RLVR reduces accuracy by 76.8\% on math tasks). 
This degradation occurs because instruction following and general reasoning (e.g., math) are often distinct capabilities; optimizing specifically for the former frequently comes at the expense of the latter. 
In contrast, Table \ref{tab:general} demonstrates that our method maintains the model's general abilities. 
This suggests that the model has acquired a foundational meta-skill that enhances instruction following without compromising its broader competencies.

\begin{figure}[t]
    \centering
    \includegraphics[width=0.8\columnwidth]{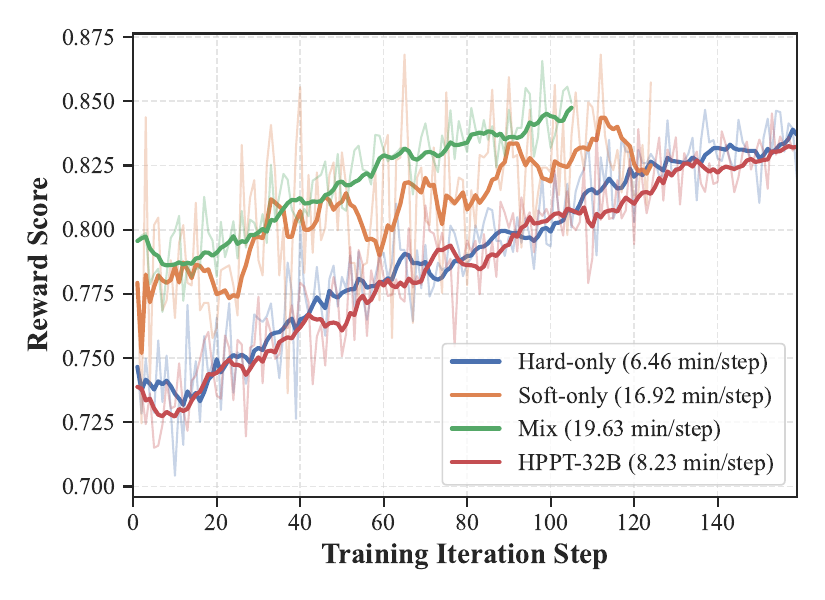}
    \caption{The training curves with computational time costs on Qwen2.5-32B-inst.
    HPPT-32B cuts overhead by 58.07\% via constraint simplification, and reward score distribution closely aligned with the \textit{Hard-only}. 
    }
    \label{fig:efficiency}
\end{figure}

\textbf{Training Efficiency Analysis.} 
Figure \ref{fig:efficiency} illustrates the training curves and computational time costs of HPPT-32B under identical experimental settings. 
Results indicate that the \textit{Hard-only} model exhibits the highest efficiency as it relies solely on rule-based verifiers, avoiding the overhead of LLM-based evaluation. 
Conversely, the \textit{Soft-only} or \textit{Mix} baselines suffer from high computational costs due to LLM reward, averaging 16.92 or 19.63 minutes per training step. 
In contrast, our method requires only 8.23 minutes per step, reducing the per-step latency by approximately 58\%. 
Furthermore, HPPT’s reward score distribution is nearly identical to the hard-only baseline (with high precision reward via rule-based checker), indicating its high-precision reward signal.
These findings demonstrate that constraint simplification strategy not only enhances reward precision but also drastically reduces the computational overhead of the verification process.

\section{Related Work}
Instruction following requires models to generate responses that satisfy complex user instructions.
Adhering to complex constraints, particularly those involving a greater number and variety of conditions, remains a challenge for LLMs ~\citep{zhang2025cfbench,tam2024let,sun2023evaluating}.
Earlier works favored sophisticated data synthesis, such as self-dialogue in AutoIF ~\citep{dong2024self}, rule extraction in RNR ~\citep{wang2024rnr}, or verifiable data generation in VFF ~\citep{wang2025verifiable}, to scale supervised fine-tuning or DPO ~\citep{kim2025systematic,jiang2024followbench,chung2024scaling,pyatkin2025generalizing}. 
The prevailing consensus, reflected in benchmarks like FollowBench ~\citep{jiang2024followbench} and CFBench ~\citep{zhang2025cfbench}, assumes that scaling constraint diversity (mixing hard and soft constraints) is essential for generalization.
This view continues to shape the emerging RLVR training paradigm ~\citep{peng2025verif}.

\textbf{Reinforcement learning with verifiable rewards.} 
This approach has attracted growing attention for its effectiveness in incentivizing reasoning in LLMs with rule-based rewards ~\citep{yue2025does,zhu2025surprising,dai2025s,2025kimik2}.
This paradigm has proven highly effective in reasoning domains, such as math and coding ~\citep{zeng2025simplerl,fatemi2025concise,liu2025deepseek}.
However, the application of RLVR has largely been confined to tasks with explicit ground truth. 
Driven by this diversity-centric view, researchers have turned to LLM-as-a-judge to provide reward signals for extending to unverifiable semantic constraints ~\citep{pyatkin2025generalizing,guo2025recast,peng2025verif,lambert2024tulu}.
In contrast, our work challenges this trend by highlighting the primacy of reward precision: we show that strict alignment with high-precision constraints alone is sufficient to cultivate robust IF meta-skills, and that excessive pursuit of diversity can actually degrade generalization.

\section{Conclusion}
This work challenges the prevailing consensus in RLVR, demonstrating that \textit{reward precision}, rather than constraint diversity, is the decisive factor for effective instruction following.
We reveal that verifiable hard constraints act as high-precision proxies, 
whereas soft constraints degrade performance due to reward hacking. 
By implementing simple {high-precision proxy training}, we achieve superior alignment performance. 
Thus, we advocate that future IF research prioritize reward engineering precision over blindly scaling constraint diversity.

\section*{Limitations}
Despite the promising results, this work presents two primary limitations that warrant further investigation:
\begin{itemize} 
    
    \item \textbf{Scalability of verifiable proxies.} Constructing programmatic verifiers (rule-based rewards) for extremely complex or abstract intents is not always feasible. 
    For domains where no clear verifiable proxy exists, the reliance on LLM-as-a-judge remains unavoidable, necessitating future work on improving the intrinsic robustness of model-based evaluators against reward hacking.

    \item \textbf{Absence of quantitative evaluation for internalized capabilities}. Our current approach relies on attention analysis as a qualitative proxy. We lack a definitive quantitative metric to precisely evaluate the internalization of the instruction-following meta-skill.

\end{itemize}

\bibliography{custom}
\appendix

\section{Contribution}
This work advocates for a paradigm shift in RLVR: moving away from the blind pursuit of data diversity toward high-precision proxy training.
Our contributions are threefold:
\begin{enumerate}
    \item Finding: identifying a counter-intuitive phenomenon in multi-constraint RLVR, the benefits of scaling constraint diversity are strictly bounded by the precision of the reward signal.
    \item Mechanism: We elucidate the hard-to-soft transfer mechanism, showing how enforcing strict adherence with verifiable rewards can generalize across unseen soft IF tasks 
    \item Method: We propose a simple yet effective data-centric refinement strategy to enable training with high-precision proxies.
\end{enumerate}

\section{Discussion and Future Works}
\textbf{Reward Precision vs. Diversity.} 
Although our results demonstrate that hard-only models with high precision yield superior generalization compared to soft-only models characterized by high diversity, this does not imply that diversity is negligible. 
On the contrary, we argue that expanding diversity is beneficial, but only when premised on high reward precision, a prerequisite likely overlooked by prior suboptimal methods.

Therefore, a promising direction for future work is to explore frameworks that simultaneously achieve high precision and high diversity to enhance robustness. 
Potential pathways include: 
(1) \textbf{Mitigating Reward Hacking}: Utilizing stronger models or training dedicated reward models to maintain precision while scaling diversity. 
(2) \textbf{Alternative Metrics}: Replacing LLM-based soft constraints with deterministic strategies. 
For example, \citet{chang2025bleuberi} propose using string-matching metrics like BLEU as an effective substitute for LLM judges.


\section{Benchmark Details}
\label{sec:benchmark}
To comprehensively assess the instruction-following capabilities of LLMs, we utilize five distinct benchmarks. 
These datasets cover hard and soft constraints, and also include generalization to unseen constraints, multi-level difficulty, and multi-turn multilingual interactions. 
Table~\ref{tab:benchmarks} summarizes the statistics of these datasets.

\textbf{IFEval} ~~\cite{zhou2023instruction} is a widely adopted benchmark designed to evaluate the ability of LLMs to follow objective and verifiable instructions. It consists of around 500 prompts containing 25 types of verifiable constraints (e.g., word count limits, formatting requirements).

\textbf{Multi-IF}~~\cite{he2024multi} extends the scope of instruction following to multi-turn and multilingual settings. It contains 4,501 samples spanning 8 languages. The dataset is constructed by expanding single-turn verifiable instructions into coherent three-turn dialogues. It serves as a stress test for maintaining instruction adherence over long contexts and across diverse linguistic distributions.

\textbf{IFBench}~~\cite{pyatkin2025generalizing} addresses the issue of model overfitting to common instruction datasets. It focuses on evaluating the generalization capabilities of models by introducing 58 novel, unseen, and challenging verifiable constraints. It employs strict code-based verification modules to measure performance on out-of-domain instructions.

\textbf{FollowBench}~~\cite{jiang2024followbench} evaluates the robustness of LLMs through a multi-level difficulty mechanism. It contains 820 instructions across five fine-grained categories (Content, Situation, Style, Format, Example). The benchmark is constructed by incrementally adding constraints to seed instructions, creating a difficulty gradient (Level 1 to Level $N$).

\textbf{CFBench}~~\cite{zhang2025cfbench} is a benchmark comprising 1,000 high-quality samples derived from the real world. It features a hierarchical taxonomy with 10 major constraint categories (e.g., style, logical rules, numerical) and over 25 subcategories. CFBench is designed to simulate complex tasks, including both an \textit{Easy} and \textit{Hard} subset to test models on varying degrees of constraint complexity.

\begin{table*}[th]
    \centering
    \small
    \begin{tabular}{l|cccc}
        \toprule
        \textbf{Dataset} & \textbf{Size (Samples)} & \textbf{Constraint Types} & \textbf{Key Features} & \textbf{Eval. Method} \\
        \midrule
            {IFEval}~~\cite{zhou2023instruction} & 500 & hard / 25 & Verifiable, Objective & Code-based \\
            {IFBench}~~\cite{pyatkin2025generalizing} & 300 & hard / 58 & Unseen Constraint & Code-based \\
            {Multi-IF}~~\cite{he2024multi} & 4,501 & hard / 25 & Multi-turn, Multilingual  & Code-based \\
            {FollowBench}~~\cite{jiang2024followbench} & 820 & mixed / 5 & Multi-level Difficulty  & Code + LLM \\
            {CFBench}~~\cite{zhang2025cfbench} & 1,000 & mixed / 25 & Complex Scenarios & LLM \\
        \bottomrule
    \end{tabular}
    \caption{Statistics and characteristics of the instruction following benchmarks.}
    \label{tab:benchmarks}
\end{table*}

\section{Detailed Experimental Results}
\label{sec:exp_details}

\subsection{GRPO as an RLVR algorithm}
\label{sec:grpo}
We employ Group Relative Policy Optimization (GRPO) ~\citep{shao2024deepseekmath} as our core learning algorithm. 
Unlike traditional PPO which relies on a value function critic, GRPO leverages group-based sampling to estimate baselines. This approach generates multiple candidate responses for the same instruction and computes advantage estimates through intra-group comparisons, effectively capturing the relative quality differences among responses.

Formally, for every input instruction $x$, the policy $\pi_\theta$ samples a group of $G$ candidate responses $\{y_i\}_{i=1}^G$. The optimization objective is defined as:
\begin{equation}
    \small
    \begin{split}
    \mathcal{J}_{\text{GRPO}}(\theta) & =  \mathbb{E}_{\substack{x \sim P(X), \\ \{y_i\}_{i=1}^G \sim \pi_{\theta_{\text{old}}}(Y|x)}} \Bigg[ \frac{1}{G} \sum_{i=1}^G  \frac{1}{|y_i|} \sum_{t=1}^{|y_i|} \bigg\{  \\
    & \min \left( \rho_{i,t} \hat{A}_{i,t}, \text{clip} \left( * \right) \hat{A}_{i,t} \right) - \beta \mathbb{D}_{\text{KL}} \left[ \pi_\theta \| \pi_{\text{ref}} \right] \bigg\} \Bigg]
    \end{split}
\end{equation}
where $\rho_{i,t} = \frac{\pi_\theta(y_{i,t} \mid x, y_{i,<t})}{\pi_{\theta_{\text{old}}}(y_{i,t} \mid x, y_{i,<t})}$ is the probability ratio. 
The \(\text{clip}(*) \) denotes \( \text{clip}\left( \rho_{i,t}, 1-\varepsilon, 1+\varepsilon \right) \).
The advantage $\hat{A}_i$ is standardized within the group to reduce variance:
\begin{equation}
\begin{aligned}
\mu &= \frac{1}{G} \sum_{i=1}^G r_i, & 
\sigma &= \sqrt{\frac{1}{G} \sum_{i=1}^G (r_i - \mu)^2 + \epsilon} \\
\hat{A}_i &= \frac{r_i - \mu}{\sigma}, & 
\rho_{i,t} &= \frac{\pi_\theta(y_{i,t} \mid x, y_{i,<t})}{\pi_{\theta_{\text{old}}}(y_{i,t} \mid x, y_{i,<t})}
\end{aligned}
\end{equation}
Note that $\epsilon$ is a small constant for numerical stability.

\subsection{Constraint Verification and Sparse Reward Design}
\label{sec:verification}

To ensure precise instruction following, we implement a dual-mode verification scheme. We classify constraints into two categories: \textit{hard constraints}, which involve objective requirements verifiable through deterministic code, and \textit{soft constraints}, which involve subjective qualities requiring model-based evaluation.
Let $C = \{c_1, \dots, c_m\}$ be the set of constraints for instruction $x$. We define the verification function $f(x, y, c_k)$ as:

\begin{equation}
f(x, y, c_k) = \begin{cases} 
V_{\text{rule}}(x, y, c_k) & \text{if } c_k \text{ is hard} \\
V_{\text{model}}(x, y, c_k) & \text{if } c_k \text{ is soft}
\end{cases}
\end{equation}

where $V_{\text{rule}}$ and $V_{\text{model}}$ denote rule-based and LLM-based validators, respectively. Both return binary outcomes ($1$ for satisfaction, $0$ for violation).

Different from previous works that utilize dense rewards (e.g., average satisfaction rate), we enforce a {strict satisfaction criterion} to prioritize precision. The final reward $R(x, y)$ is binary, granted only when \textit{all} constraints are simultaneously satisfied:
\begin{equation}
R(x, y) = \prod_{k=1}^{|C|} f(x, y, c_k) = \mathbb{I}\left( \forall c_k \in C, c_k \text{ pass} \right)
\end{equation}
This sparse reward signal drives the model to strictly adhere to the complete set of instructions, penalizing any partial failure.

\subsection{Implementation Details}
For reinforcement learning, we implemented GRPO based on the MindSpeed-RL\footnote{https://gitcode.com/Ascend/MindSpeed-RL} training framework. 
Each RL training run for the 7B model completed within 24 hours on a cluster of 64 Ascend 910b NPUs (configured as 8 nodes × 8 NPUs). 
For optimization stability, we incorporated KL divergence regularization with a coefficient of 0.001 using the low-variance KL implementation, while enabling gradient checkpointing for memory efficiency. 
The hyperparameters used are detailed in Table \ref{tab:grpo_config}.
\begin{table}[ht]
    \centering
    \begin{tabular}{l|c}
        \toprule
        \textbf{Hyperparameter} & \textbf{Value} \\
        \midrule
        \multicolumn{2}{c}{\textit{Data Configuration}} \\
        \midrule
        Global Batch Size & 128 \\
        Max Prompt Length & 2048 \\
        Max Response Length & 6144 \\
        Micro Batch Size & 4 \\
        Train Steps & 600 \\
        \midrule
        \multicolumn{2}{c}{\textit{Rollout Configuration}} \\
        \midrule
        Rollout Name & vllm \\
        GPU Memory Utilization & 0.6 \\
        Number of Rollouts & 8 \\
        Temperature & 1.0 \\
        Tensor Model Parallel Size & 4 \\
        Top\_P & 1.0 \\
        \midrule
        \multicolumn{2}{c}{\textit{RL Optimization}} \\
        \midrule
        Learning Rate & 1e-6 \\
        LR Decay Style & constant \\
        Mini Batch Size & 128 \\
        KL Loss & 0.001 \\
        \bottomrule
    \end{tabular}
    \caption{The configurations for RL training with GRPO.}
    \label{tab:grpo_config}
\end{table}

\begin{table*}[t]
    \centering
    \small
    \begin{tabular}{l|ccc|cc|c}
        \toprule
         Model & \ding{168}IFEval & \ding{168}Multi-IF & \ding{168}IFBench & \ding{171}CFBench & \ding{171}FollowBench & \textbf{Average} \\
        \midrule 
        \midrule \rowcolor{lightgray!40}
        Qwen2.5-7B-inst & 72.46 & 51.05 & 28.91 & 44.00 & 61.40 & 51.56 \\ 
        \hspace{0.2cm} \textit{w/ {step a}} & 85.03\up{12.6} & 65.69\up{14.6} & 38.09\up{9.2} & 54.00\up{10.0} & 70.83\up{9.4} & 62.73\up{11.2} \\
        \rowcolor{blue!10} \hspace{0.2cm} \textit{w/ step a+b} & 87.25\up{14.8} & 68.70\up{17.7} & 40.13\up{11.2} & 57.00\up{13.0} & 70.88\up{9.5} & 64.79\up{13.2} \\
        \bottomrule
    \end{tabular}
    \caption{Ablation study of HPPT-7B. We observe that step (a) learnability filtering contributes the most significant improvement, while step (b) constraint simplification further enhances performance across all benchmarks. 
    }
    \label{tab:ablation}
\end{table*}

\subsection{Baseline RLVR-trained Models}
\label{sec:baseline}
We conduct comparisons against three representative recent RLVR-trained models to improve the instruction-following capabilities of language models, namely:
\begin{itemize}
    \item RECAST ~\citep{guo2025recast}, a framework that empowers models to handle complex, multi-constraint instructions by utilizing a verifiable data synthesis pipeline and RL with verifiable constraints.
    \item IF-RLVR ~\citep{pyatkin2025generalizing}, training RLVR with multi-constraints per instance to enhance a model's ability to follow diverse and complex hard constraint instructions.
    \item Qwen-IF ~\citep{ren2025instructions}, a label-free self-supervised RL method that enhances IF by deriving reward signals directly from input instructions and utilizing constraint decomposition to address sparse rewards.
\end{itemize}

\section{Details in Reward Reliability}
\label{sec:human_label}

\paragraph{Ground-Truth Curation Protocol.}
To construct a robust benchmark for validating automated reward signals, we curated a representative subset comprising $N=200$ instances, balanced equally between verifiable hard-constraint tasks (sampled from IFEval) and semantic soft-constraint tasks (sampled from CFBench).
We recruited three domain experts for blind evaluation.
To rigorously quantify label reliability, we calculated Fleiss’ Kappa ($\kappa$) to measure Inter-Annotator Agreement (IAA).
We observed near-perfect consistency for hard constraints ($\kappa = 0.91$) and substantial agreement for soft constraints ($\kappa = 0.78$).
For instances exhibiting significant divergence (defined as a variance $> 0.8$ on the Likert scale or conflicting binary verdicts), a designated senior lead researcher served as an adjudicator to determine the final ground truth.
Consequently, the benchmark scores are derived from this adjudicated consensus rather than simple averaging, ensuring a high-confidence gold standard.

\paragraph{Inference Stability and Recovery.}
Evaluating instructions with complex constraints requires robust inference pipelines.
During batch-wise evaluation, we identified edge cases where the model exhibited stochastic degradation, such as repetition loops or degenerate token sequences.
To guarantee the validity of the reward signals, we integrated an automated sanity check and recovery module.
If an output is detected as malformed or unparsable, the system triggers an iterative resampling process with a non-zero temperature, continuing until a compliant response is acquired or the retry budget is exhausted.


\section{Ablation Study}
\label{sec:ablation}
We conduct an ablation study to analyze the effectiveness of our HPPT-7B model by isolating two key components: 
(a) Denoising via learnability filtering; 
(b) Mitigating hacking via constraint simplification. 
Following the same experimental setup as Section~\ref{sec:method}, we present the results in Table~\ref{tab:ablation}.
The results indicate that both steps are indispensable. 
Specifically, applying learnability filtering (step a) yields a substantial average improvement of 11.2\% over the baseline. 
Incorporating constraint simplification (step b) further boosts the performance, resulting in a total average gain of 13.4\%.

\end{document}